\documentclass[journal]{IEEEtran}

\usepackage{threeparttable}
\usepackage{booktabs}
\usepackage{graphicx}
\usepackage{amsfonts}
\usepackage{hyperref}

\begin{document}

\title{On the Learnability of Footwear Robustness: Footstep Recognition Under Unseen Conditions}

\author{Robyn~Larracy,~\IEEEmembership{Graduate Student Member,~IEEE,}
        and~Erik~Scheme,~\IEEEmembership{Senior~Member,~IEEE}
\thanks{R. Larracy and E. Scheme are with the Movement and Biosignals Intelligence Lab (MoBIL), University of New Brunswick, Fredericton, NB, Canada. e-mail: \{rlarracy,escheme\}@unb.ca.}}

\markboth{Submitted to IEEE TRANSACTIONS ON INFORMATION FORENSICS AND SECURITY}%
{Larracy \MakeLowercase{\textit{et al.}}: On the Learnability of Footwear Robustness}

\maketitle

\begin{abstract}

Footstep patterns, captured using pressure or force-sensitive walkways during natural walking, offer a convenient and passive approach to biometric recognition. 
However, footwear variability poses a major challenge to practical deployment, as it impacts both the foot-sensor interface and the user's underlying gait mechanics. 
This study considers the particularly difficult scenario where users enroll under one fixed footwear condition but must later be recognized in arbitrary, unseen footwear.
Using the StepUP-P150 dataset, we show that footwear-induced shift, while substantial and heterogeneous across individuals, contains shared structure that can be learned from a training population and transferred, zero-shot, to previously unseen users and footwear styles.
We do this using BiStepNet, a novel architecture that decomposes spatial and temporal identity cues into independently trainable streams and outperforms compact 3D spatiotemporal baselines for cross-footwear recognition. 
We demonstrate that cross-footwear recognition performance scales systematically with the volume and diversity of footwear representation in the training set, with gains transferring to footwear styles not directly observed during training. 
Moreover, we show that auxiliary gait datasets with only partial modality overlap (incorporated through the architecture's decomposed design) can be leveraged to further improve robustness to unseen footwear conditions.
\end{abstract}


\IEEEpeerreviewmaketitle

\section{Introduction}
%
%
%


\IEEEPARstart{F}{ootstep} recognition is the identification or verification of individuals from their ground-contact signals generated during walking. 
Sensors embedded in floors or walkways can passively and continuously capture these signals, in the form of plantar pressure distributions or ground reaction forces, without requiring users to pause, present themselves, or remove equipment or personal items for capture. 
Moreover, unlike vision-based gait recognition, these sensors measure the biomechanical signals underlying gait directly rather than inferring them from video, making them inherently robust to challenges such as occlusion and viewpoint variation.
These properties make footstep recognition a convenient option for physical access control and activity monitoring in environments like airports, secured offices, and other restricted facilities. 
Recent studies have demonstrated that footstep recognition can achieve strong performance under constrained and familiar conditions, with reported identification and verification accuracies routinely exceeding 95\% in barefoot or fixed-footwear settings \cite{Wei2026,Zhang2023,Horst2023,Derlatka2023}.

Real-world deployment, however, introduces covariates that can substantially degrade this performance, with footwear posing the most significant challenge.
Although some applications naturally limit footwear variation (e.g., airport security checkpoints requiring shoe removal, workplaces with mandated footwear, or short-term recognition scenarios where footwear is unchanged), many practical deployments must accommodate footwear changes between enrollment and subsequent transactions.
Footwear affects the recorded signals in two compounding ways: first, by altering the foot's interface with the sensor, redistributing and filtering the plantar pressures, and second, by modifying the user's underlying gait mechanics.
In some cases, these effects can be large enough to overwhelm the biometric information in the pressure signal.
Zhou et al. \cite{Zhou2026} demonstrated that within-person differences across footwear conditions exceeded between-person differences within the same footwear condition, and studies consistently report large drops in performance for cross-shoe recognition relative to within-footwear recognition \cite{Derlatka2017,Duncanson2023,Zhou2026}.

This study considers a particularly challenging cross-shoe recognition scenario, where users enroll under a single, fixed footwear condition (barefoot) but may present in arbitrary, unconstrained footwear during subsequent transactions. 
In this setting, footwear-robust representations generally cannot be learned from a user's own enrollment data, since 1) only one condition is available per user, and 2) retraining or adapting the model for each new enrollee is impractical at deployment scale. 
Instead, these representations must be learned from a separate set of training identities and generalize, zero-shot, to previously unseen users wearing (potentially) previously unseen footwear styles. 
While a handful of studies have explored recognition across footwear conditions, these often assume either closed-set evaluation protocols \cite{Zhou2026,Connor2015,Derlatka2017} or a small set of predefined footwear domains \cite{Wang2024disencb}. 
Consequently, it remains unclear whether footwear-robust representations learned from one group of individuals can transfer to new users and new styles alike. 

Using the StepUP-P150 dataset \cite{Larracy2025}, which includes high-resolution footstep pressure recordings for 150 individuals under four footwear conditions each, we investigate whether transferable footwear-robust representations can be learned for zero-shot cross-footwear recognition.
To answer this question, we proceed in three parts:

\begin{enumerate}
    \item We introduce \textbf{BiStepNet} (\textbf{Bi}-stream \textbf{S}patial-\textbf{Te}mporal \textbf{P}ressure \textbf{Net}work) which separately encodes spatial and temporal features using established gait representations, and show that this decomposed architecture outperforms several compact 3D spatiotemporal baselines.
    \item We characterize the relationship between footwear representation during training and cross-footwear recognition performance, finding that performance improves systematically with additional footwear exposure and extends to footwear styles not directly represented during training.
    \item We propose a multi-dataset contrastive training strategy that leverages the decomposed network design to incorporate auxiliary gait datasets with partial modality coverage, and show meaningful gains from the added diversity of identities and conditions.
\end{enumerate}


\section{Related Work}
As an emerging field, much of the early footstep recognition literature has focused on validating the uniqueness of footstep recordings by training classifiers to recognize a fixed set of users under constrained conditions. 
Most of these studies evaluate participants walking either barefoot \cite{Zhang2023,Horst2023,Pataky2012,Zheng2012} or in a single pair of their own personal footwear \cite{Wei2026,Derlatka2023,Terrier2020,Majeed2021} (a protocol which, given the strong influence of footwear, risks encouraging models to overfit to shoe-specific patterns rather than underlying identity features). 
By contrast, only a handful of studies have evaluated cross-footwear recognition or proposed techniques to improve robustness to footwear variation.

Connor \cite{Connor2015} and Zhou et al. \cite{Zhou2026} approached cross-footwear recognition by exploiting long-range temporal information and gait attributes, such as stride length and width, extracted from several consecutive footsteps during inference. 
Connor employed a hand-crafted feature pipeline that combined spatial and temporal footstep features with gait-level descriptors, followed by principal component analysis (PCA) and forward feature selection. 
Using five consecutive footsteps, the method achieved a cross-shoe rank-1 identification accuracy of 99.5\% and a verification equal error rate (EER) of 11.4\% on the 13-user CASIA-D dataset. 
Zhou et al. proposed a transformer architecture that incorporated physical and gait attributes as auxiliary prompts. 
Evaluated on the 200-user MFID-200 dataset, the model achieved 94.61\% within-footwear (barefoot-to-barefoot) and 73.42\% cross-footwear (barefoot-to-mixed-footwear) rank-1 accuracy using sequences of approximately 14 consecutive footsteps. 
However, both studies employed subject-dependent (closed-set) evaluation protocols in which the same identities were used for training and testing. 
Consequently, neither investigated whether the learned representations generalize to previously unseen users.

Wang et al. \cite{Wang2024disencb}, by contrast, explicitly evaluated generalization to unseen identities by training with 140 users and testing with 60 held-out users. 
They proposed a domain-adversarial disentanglement framework that fused domain-relevant and domain-irrelevant features, achieving a cross-footwear rank-1 identification accuracy of 77.78\%. 
However, the evaluation was restricted to two standardized footwear conditions (barefoot and a cloth shoe) that were shared across all participants. 
Although this setting isolates the effect of a controlled footwear change, it assumes a small, fixed set of common domains that is unlikely to represent real-world deployment, where shoes can vary continuously in style and construction. 

To our knowledge, only one prior study considers the current scenario of cross-footwear recognition against a probe set of unseen identities and varied, possibly unseen footwear styles. 
Duncanson et al. \cite{Duncanson2023} collected a force-plate dataset of 193 participants who each returned for two sessions, with 78 wearing different personal footwear between sessions. 
Using seven-fold cross-validation, they trained several deep network architectures with a triplet-loss metric learning approach and evaluated performance on held-out identities. 
Averaged across five architectures, within-footwear rank-1 accuracy ranged from 81.3\% to 95.7\%, while cross-footwear accuracy ranged from 68.3\% to 83.2\%, an average drop of 14 percentage points.

Our work adopts a similar contrastive metric-learning framework for zero-shot cross-footwear recognition, but differs in several important respects. 
First, rather than relying on force plate recordings, we use a high-density pressure sensing modality that captures both spatial and temporal characteristics of foot-ground interaction. 
Second, we evaluate on the StepUP-P150 dataset \cite{Larracy2025}, which includes four footwear conditions per participant, including two standardized conditions shared across all participants. 
This controlled diversity enables systematic investigation of how footwear representation during training influences cross-footwear generalization, rather than relying on incidental variation in personal footwear. 
Building on this foundation, we investigate whether footwear-induced variability contains sufficient shared structure for representations learned from one population to generalize to previously unseen users and footwear conditions. To answer this question, we propose a robust spatiotemporal architecture with an auxiliary multi-dataset training strategy designed to improve robustness to unseen footwear conditions.

\section{Footwear-Induced Domain Shift}\label{sec:problem_formulation}

Changes in footwear introduce a domain shift in the recorded pressure data. Every footwear condition, including barefoot walking, can therefore be regarded as a distinct domain. 
During deployment, this manifests as a reference-probe mismatch, where the footwear worn during a verification or identification transaction differs from that worn by the claimed user during enrollment.
We target an especially challenging version of this problem, in which a user enrolls with a small number of footsteps in a single footwear condition and must later be recognized while wearing different, arbitrary styles. 
Because footwear-robust representations cannot be learned from a user's own enrollment data, they must instead be learned from a separate pool of training identities and generalized to new users at deployment.
Formally, these shoe-induced effects constitute a class-conditional shift, in which the conditional measurement distribution $P(X|Y)$ varies across footwear domains.
The objective is therefore to learn an encoder $\phi$ that produces representations for which $P(\phi(X)|Y)$ remains invariant to footwear, even for identities ($Y = y$) that were unseen during training. 

This setup presents several challenges. 
First, the space of possible shoe styles a system may encounter during deployment is effectively unbounded, so the model will inevitably be faced with novel domains not represented during training.
Moreover, training sets will generally have disjoint label sets across domains.
For example, one subset of users might wear large winter boots, another high heels, with little or no overlap between groups. 
As a result, unless common footwear domains are specifically collected (as with the StepUP-P150 \cite{Larracy2025} and MFID-200 \cite{Zhou2026} datasets), the model has limited opportunity to observe how a given domain shift manifests across multiple individuals. 
Shoe style domains are also themselves ill-defined, since continuous variation in characteristics such as heel height and sole softness produces a spectrum of effects rather than discrete categories, and coarse labels like running shoes, sandals, or casual sneakers are subjective and may overlap.
Rather than relying on fine-grained domain labels, it is therefore more tractable to exploit grouped observations during training using multiple examples from the same identity collected under different, unlabelled conditions. 
This motivates a contrastive or metric-learning approach over explicit domain alignment, disentanglement, or translation-based methods. 

\begin{figure}
    \centering
    \includegraphics[width=\linewidth]{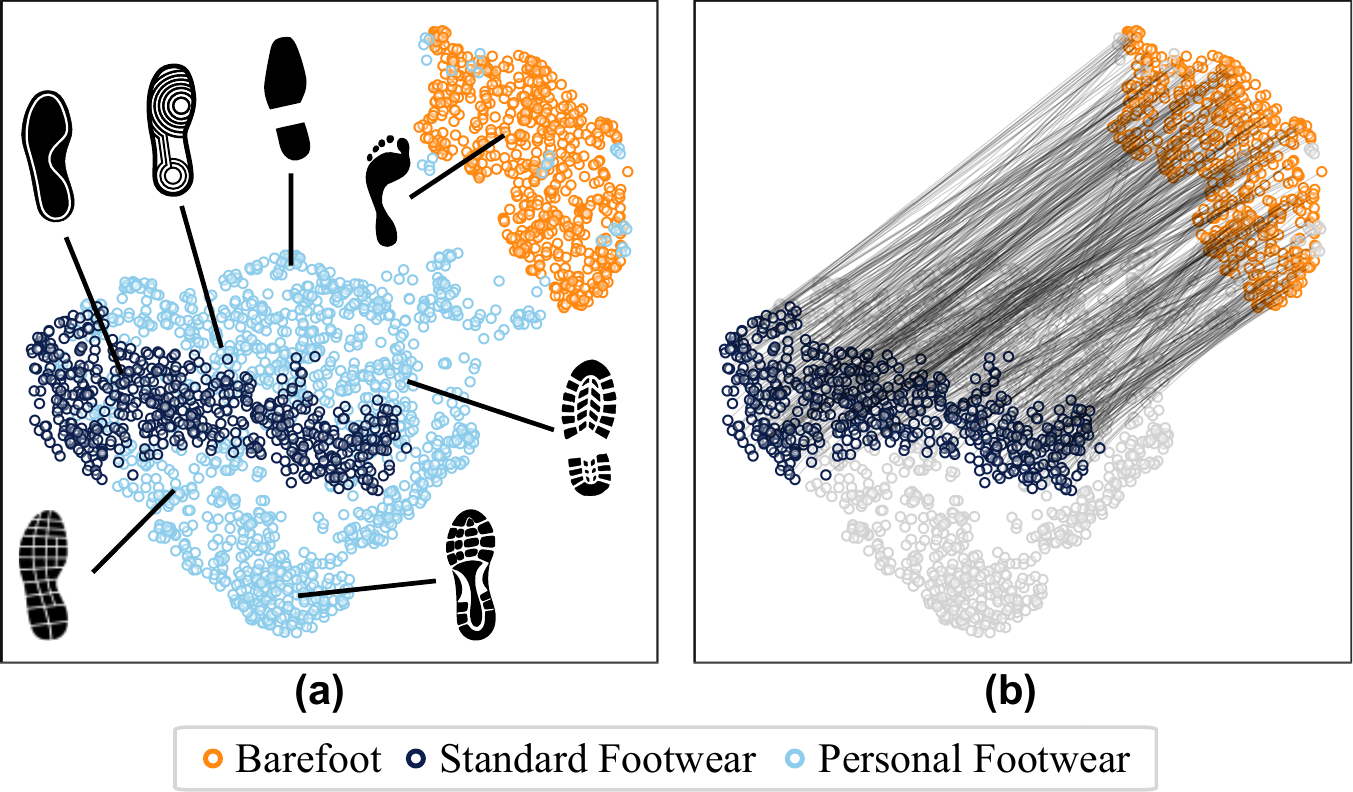}
    \caption{Demonstration of the domain shift between pressure recordings collected in different footwear conditions: (a) footwear clusters in UMAP space, (b) per-user shift vectors between walking barefoot and walking in a common, standardized shoe (Adidas casual sneakers). The shift has a dominant shared direction across individuals.}
    \label{fig:domain_shift}
\end{figure}

Lastly, although literature suggests that certain shoe characteristics will have similar effects across people \cite{Mohammadi2025}, the domain shift cannot be expected to be homogeneous.
As an illustration of this heterogeneity, we project the average footstep recording from each trial in StepUP-P150 via PCA (retaining components explaining 95\% of the variance) and visualize the shift between barefoot and a shared standard shoe for 150 individuals using UMAP (Fig. \ref{fig:domain_shift}). 
This reveals a dominant shift direction shared across participants (mean pairwise cosine similarity 0.64, vs. 0.42 for arbitrary personal footwear), alongside a substantial participant-specific residual (on average, 73\% the magnitude of the population component), indicating that shoe-induced effects are neither fully uniform nor fully idiosyncratic. 

This structure underlies the central challenge of cross-shoe recognition. A shift with meaningful shared structure suggests that shoe-invariant features should, in principle, be learnable, while the heterogeneous components suggest that doing so will require sufficient identity diversity during training to capture how footwear characteristics will manifest for any individual the system might encounter.
We hypothesize that, given adequate identity and footwear diversity during training, this structure is learnable and transferable to previously unseen users, and explore this hypothesis empirically in this study.




\begin{figure*}
    \centering
    \includegraphics[width=\linewidth]{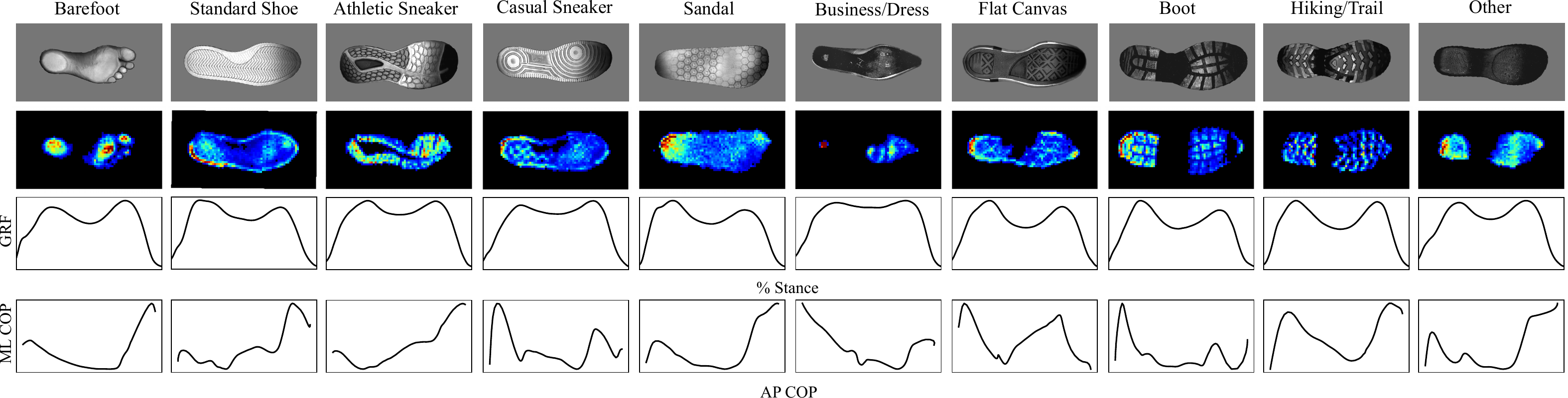}
    \caption{Examples of footwear conditions in the StepUP-P150 dataset, with corresponding peak pressure images, ground reaction force (GRF) time series, and center of pressure (COP) trajectories in the mediolateral (ML) and anteroposterior (AP) directions. The conditions are shown in order of their prevalence in the dataset; barefoot and standard shoe conditions were collected for all 150 participants, while the remaining categories included personal shoe styles that varied across participants.  
    Note: GRF and COP are shown in normalized units to facilitate visual comparison across conditions.} 
    \label{fig:example_shoes}
\end{figure*}

\section{Methods}

\subsection{Experimental Setup}
\subsubsection{Primary Dataset}
The publicly-available StepUP-P150 dataset was the primary dataset used for the experiments \cite{Larracy2025}. 
This dataset contains footstep pressure recordings from 150 individuals (74 M, 76 F; 19–91 years) as they walked across a 1.2 m \texttimes~3.6 m instrumented flooring system equipped with high-density pressure sensors (4 pixels/cm$^2$) sampled at 100 Hz.
Participants completed a total of sixteen 90-second walking trials, including all combinations of four walking speeds (preferred, slow, fast, slow-to-stop) and four footwear conditions (barefoot or sock-foot, standard Adidas sneakers, and two pairs of the participant's personal shoes). 
These personal shoes included a diverse range of footwear, coarsely categorized into eight style families; see example styles and corresponding footstep recordings in Fig. \ref{fig:example_shoes}.
All participants provided their informed written consent, and the study was approved by the University of New Brunswick's Research Ethics Board (REB 2022-132).

To process the footsteps into a format suitable for deep learning, individual footsteps were extracted from the 90-second recordings and normalized in size. 
Full details of the preprocessing steps are provided in the original paper describing the data (see preprocessing Pipeline 1 \cite{Larracy2025}); briefly, the footsteps were rotated to align their first principal component axis with the $y$-axis, spatially zero-padded to a height of 75 pixels and width of 40 pixels, and normalized in time to 101 frames.
Incomplete or outlier footsteps, identified through a combination of automated and manual review, were excluded from the analysis.

\subsubsection{Data Splitting}
Disjoint sets of individuals were used for model development and evaluation. 
A three-fold cross-validation protocol was used, with 100 individuals allocated to model development and 50 reserved for downstream biometric matching in each fold. 
Within the development set, 90 individuals were used for training and 10 for validation.  
Each input sample consisted of a complete gait cycle stride consisting of a pair of consecutive left and right footsteps.
Training samples were constructed using overlapping strides (i.e., one footstep could appear in two separate strides) to maximize data utilization, whereas validation and test samples consisted exclusively of non-overlapping strides.
On average, each fold contained approximately 92,000 development strides and 27,000 test strides.

\subsubsection{Verification and Identification}
Verification and identification performance were evaluated using a distance-based matching protocol on the 50 held-out test users in each fold. 
For each test user, five reference strides were sampled from their barefoot walking trials collected at their preferred walking speed. 
Their remaining samples, including additional barefoot strides and strides collected under three footwear conditions, served as mated probes. 
Samples from the other 49 test users in each fold served as non-mated probes; because these individuals were excluded from model development, non-mated comparisons represent matching against entirely unseen distractor identities.

This protocol simulates a deployment context in which all users perform their enrollment sessions barefoot, but footwear is unconstrained for subsequent transactions.
We therefore consider three evaluation scenarios of increasing difficulty, depending on the footwear worn for the probe stride; (1) barefoot-to-barefoot comparisons (BF~\textrightarrow~BF), (2) barefoot-to-standard-footwear comparisons (BF~\textrightarrow~SF), and (3) barefoot-to-personal-footwear comparisons (BF~\textrightarrow~PF). 
While the standard footwear style was represented during network training for other identities, the test users' personal footwear styles may or may not have been encountered at all during training. 

For scoring, all left-right footstep pairs were passed through the trained feature extraction network to obtain 256-dimensional feature embeddings. 
A prototype reference embedding was constructed for each user by averaging the L2-normalized embeddings of their five reference strides. 
Probe embeddings were then compared against the prototypes using cosine similarity to generate a score for each probe. 
Exhaustive prototype-probe comparisons were performed for each fold, yielding over 4 million comparisons in total. 
These were used for computing verification performance (equal error rate, EER) and identification performance (rank-1 accuracy). 

\subsection{Part 1: Spatiotemporal Architecture Comparison}\label{sec:StageI}

\subsubsection{Spatial and Temporal Representations}\label{sec:representations}
Because footwear is expected to affect spatial and temporal gait characteristics differently, we represent and learn these information sources independently. 
The resulting BiStepNet architecture encodes spatial and temporal information separately,  using established lower-dimensional representations from the gait analysis and recognition literature. 

Three time series representations that are widely used in force plate-based gait analysis were considered: the vertical ground reaction force (GRF) waveform and the mediolateral (ML) and anteroposterior (AP) center of pressure (COP) trajectories. 
Although force plates directly measure three-dimensional GRFs, only the vertical component was considered for this work as its shape can be reliably estimated from the pressure recordings by summing sensor values at each time step. 
Likewise, the COP, which captures the spatial progression of the load during the stance, can be computed from the pressure recordings as the pressure-weighted centroid of the impression at each time step.

Five two-dimensional representations derived from the pressure maps were used for learning spatial features. 
These included peak pressure, mean pressure, and mean contact area images, which have previously demonstrated strong performance for footstep recognition \cite{Pataky2012,Larracy2023}. 
Additionally, two representations based on first-order temporal differences were used: the maximum positive and negative differences of each pixel, capturing peak loading and offloading rates.

Prior to training, GRF waveforms were scaled by their mean amplitude and COP trajectories were mean-centered and normalized by their anteroposterior span (approximating foot length).
Likewise, all of the image representations were individually rescaled to amplitudes in the range $[0, 1]$.
For all images except the mean contact area representation, values above the 99th percentile were clipped prior to rescaling to reduce outlier effects, and a natural logarithm transformation was applied to enhance low-pressure features.

\begin{figure*}[tb]
    \centering
    \includegraphics[width=\linewidth]{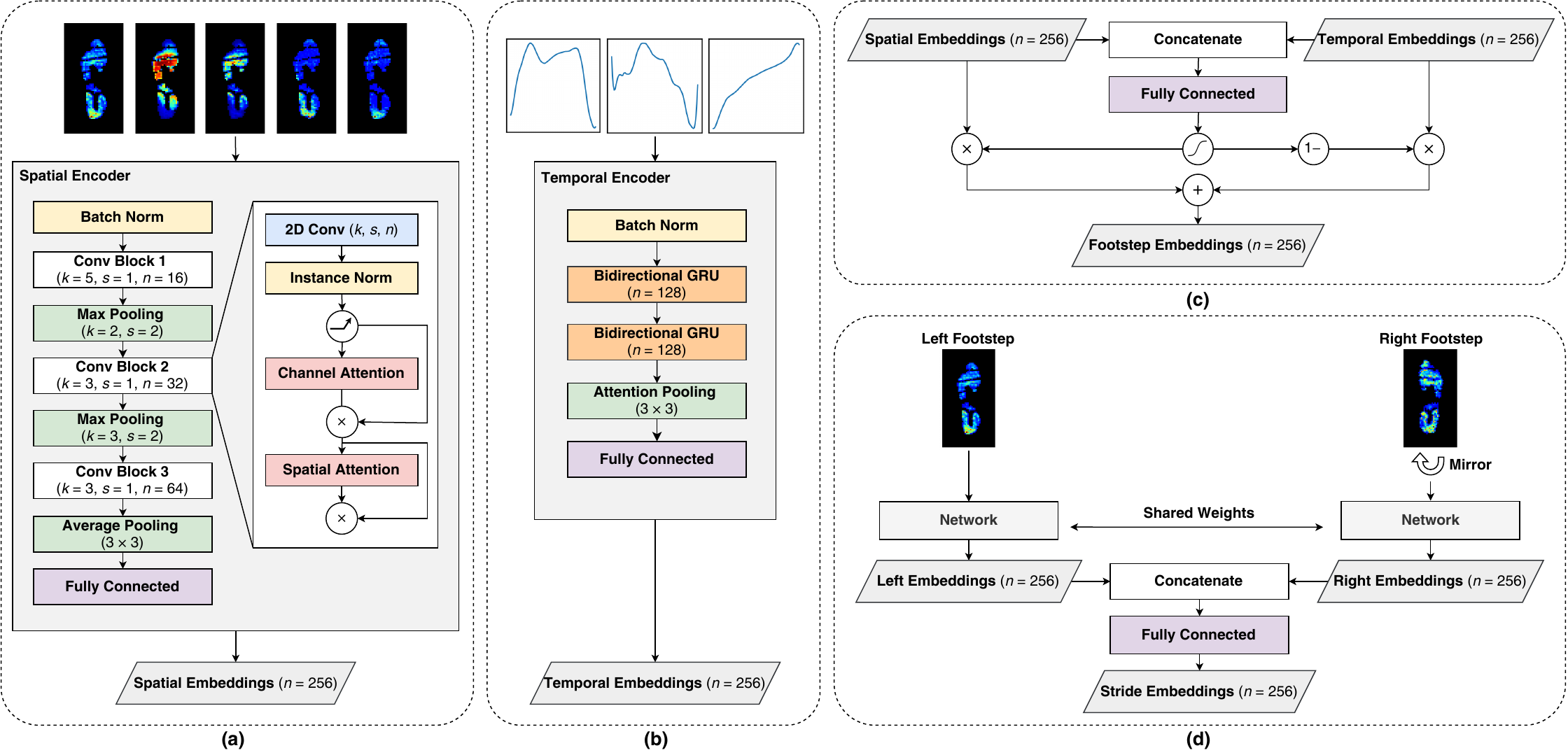}
    \caption{Overview of the proposed BiStepNet: (a) CNN-based architecture that encodes five 2D input representations; (b) GRU-based architecture that encodes three 1D input representations; (c) gated fusion module used to merge the embeddings from the spatial and temporal streams into a single embedding vector for the footstep; and (d) feature-level fusion strategy used to combine information from paired left/right footsteps into a final 256-feature vector.}
    \label{fig:network}
\end{figure*}

\subsubsection{Proposed Bi-Stream Spatiotemporal Network}
The proposed BiStepNet for feature extraction consists of two lightweight encoders; a convolutional network to encode spatial information from the 2D gait representations, and a recurrent network to encode temporal information from the 1D time series representations. 
The network architecture is shown in Fig. \ref{fig:network}.
Briefly, the spatial encoder is comprised of three convolutional blocks, which contain a 2D convolutional layer followed by instance normalization \cite{Ulyanov2016}, ReLU activation, and channel and spatial attention through a Convolutional Block Attention Module (CBAM) \cite{Woo2018}. 
Each convolutional block is followed by a pooling layer; max pooling after the first two blocks and average pooling after the third block. 
Finally, a fully connected layer produces the 256-dimensional feature vector for the stream. 
The temporal encoder consisted of two bidirectional gated recurrent unit (GRU) layers, followed by attention pooling and a fully connected projection layer to create a 256-dimensional temporal embedding. 
A gated fusion technique was used to combine the embeddings from the two streams, using a learned mapping to obtain a scalar weighting for the summation of the two vectors.
The architecture is compact, with a total of 0.6M trainable parameters.

\subsubsection{Baseline 3D Architectures}
Three spatiotemporal architectures, representative of distinct modeling families commonly used for related spatiotemporal problems, were evaluated as baselines. 
Unlike the proposed bi-stream network, these models operate directly on the full 3D footstep volume (time~\texttimes~height~\texttimes~width), learning spatial and temporal patterns jointly rather than through decomposed streams. 
These included: (1) a 3D CNN \cite{Tran2015} with three layers of 3D convolutions (0.4M parameters); (2) an R(2+1)D CNN \cite{Tran2018}, using three residual blocks with factorized spatial and temporal convolutions (0.4M parameters); and (3) a hybrid 3D CNN-GRU network, which consists of a three-block 3D CNN followed by a two-layer bidirectional GRU (0.6M parameters).
Where possible, these architectures were designed with depths and hyperparameters (e.g., kernel sizes) comparable to the proposed network. 
Full architectural details are provided in the Supplementary Material, and PyTorch implementations of all models (including BiStepNet) are available on GitHub\footnote{\href{https://github.com/rlarracy/bistepnet}{github.com/rlarracy/bistepnet}}.

Similar to BiStepNet, individual footstep recordings were normalized before training by clipping pressures above the 99th percentile, rescaling to the range $[0,1]$, and applying a natural logarithm transformation.


\subsubsection{Left and Right Footstep Fusion}\label{sec:lr_fusion}
All of the compared architectures were designed to operate on individual footsteps.
Figure \ref{fig:network}(d) shows the feature-level fusion strategy used to combine information from paired left and right footsteps in a stride. 
Each footstep was independently processed by the same backbone, and their corresponding embeddings were concatenated and passed through a fully-connected layer to obtain a fused 256-dimensional representation. 
To ensure a unified representation across sides, right footsteps were mirrored before feature extraction. 
The models were trained end-to-end with the fusion head, using paired footsteps as inputs.

\subsubsection{Training Details}
Models were trained using supervised contrastive loss ($t = 0.05$) \cite{Khosla2020} and an AdamW optimizer. 
An initial learning rate of $1\times10^{-3}$ was used following a 100-step linear warm-up, with a weight decay of $1\times10^{-6}$. 
Batches were constructed using domain-aware sampling to ensure shoe diversity at each training step, with each comprising 64 individuals \texttimes~4 shoe conditions \texttimes~2 samples, for a total batch size of 512 samples.
During training, affine augmentations were applied with a probability of 50\% to simulate minor spatial misalignments. 
These consisted of random rotations in the range $[-15^\circ, 15^\circ]$ and translations up to 10\% of the spatial dimensions along the $x$ and $y$ axes.
Validation performance was evaluated every 100 batches (treated as one epoch for scheduling purposes), computed as the area under the detection error tradeoff curve (DET-AUC) over all pairwise comparisons between validation samples. 
Learning rate was reduced by a factor of 0.5 upon plateau in validation performance (patience of 1 epoch), and training was stopped early after 5 epochs without improvement. 
The checkpoint with the best validation AUC was retained.
During both validation and testing inference, an exponential moving average (EMA) of model weights with a decay of 0.99 was used \cite{Brotons2024}, resulting in an effective memory window of approximately 100 training steps.
Data loaders were seeded identically across all training runs to ensure reproducibility and a fair comparison between models.

For the BiStepNet, we additionally considered a staged training procedure in which the two streams were first trained independently before being jointly fine-tuned, in contrast to standard combined training. 
The spatial and temporal encoders were pretrained using the hyperparameters described above, using the left/right fusion head from Section \ref{sec:lr_fusion} to accommodate paired steps. 
Following pretraining, the fusion heads were discarded and the two encoders combined for joint fine-tuning. The backbones were frozen for the first epoch, during which only the stream-gating and left/right fusion head were trained, then the backbones were unfrozen and fine-tuned together at a halved learning rate ($5\times10^{-4}$) until validation performance stopped improving. 
A dropout rate of 20\% was applied between GRU layers and at the fully connected output layer of each stream during this fine-tuning stage.

The experiments were run on a single NVIDIA RTX A4000 GPU (16 GB memory). 
All three of the baseline architectures exceeded the GPU memory budget with the 512-sample batch sizes; GradCache \cite{Gao2021} was used in these cases to enable supervised contrastive learning within the available memory. 




\subsection{Part 2: Effect of Training Set Footwear Representation}

Section \ref{sec:problem_formulation} identified a shift direction shared across participants for a given shoe style, alongside a smaller participant-specific residual. 
Building on this, we used a learning curve procedure to test whether increased footwear representation during training produces predictable gains in cross-footwear recognition performance. 

Using BiStepNet with the staged training procedure, we compared two training set expansion strategies, holding participant count fixed while varying how additional footwear representation was introduced. 
In both cases, the baseline training set consisted of all strides from two footwear domains: barefoot, and one personal footwear style per participant.  
The strategies differed only in how footwear instances beyond this baseline were added. 
Single-domain expansion incrementally added instances of a shared style (standard footwear) across participants, while multi-domain expansion instead added a second, distinct personal footwear instance per participant. 

For single-domain expansion, the number of participants contributing standard shoe recordings during training was systematically increased ($k \in \{0, 1, 2, 5, 10, 20, 50, 90\}$, where $k = 90$ corresponds to the full training partition), with recordings included for a random subset of $k$ participants at each step. 
Multi-domain expansion followed a similar procedure, by increasing the number of participants contributing training strides in a second pair of personal footwear.
For both, the random sampling procedure was repeated five times for each cross-validation fold, and performance was averaged across all folds and repetitions to produce the final learning curves.

\subsection{Part 3: Multi-Dataset Training}

\subsubsection{Auxiliary Datasets}
Five additional gait datasets were incorporated during training (Table \ref{tab:datasets}). 
Four were collected using walkways embedded with two sequential force plates to independently capture the left and right footsteps in a gait cycle, providing accurate 3D GRFs, 2D COP, and 3D orthogonal moment components during walking \cite{Horsak2020,Horst2021,DerlatkaParfieniuk2023,Duncanson2023}.
The three time series of interest (vertical GRF, 2D COP) from these datasets were used for auxiliary training of the temporal encoder. 

The fifth dataset, MFID-200, consists of high-resolution 2D peak pressure maps collected from a 10 m pressure runway \cite{Zhou2026}. 
Individual footsteps were segmented from the raw recordings, automatically labelled as left or right instances \cite{Macdonald2023}, and grouped into overlapping pairs for auxiliary training of the spatial stream. 
As MFID-200 contains only peak pressure maps, the remaining spatial input channels were approximated by using a binary pressure mask in place of mean contact area and replicating the peak pressure image across the remaining channels.

The datasets vary in footwear coverage, some including one or multiple personal or standardized footwear conditions per user, while one is barefoot only.
Some also include additional sources of variability such as varied walking speeds, loading conditions, or cross-day collection. 
All datasets were preprocessed to approximately match the resolution and format of StepUP-P150, and the same normalization procedures (described in Section \ref{sec:representations}) were used for all datasets to improve consistency during pooled training.
Full preprocessing details are provided in the Supplementary Material.

\begin{table}[t]
\centering
\caption{Summary of auxiliary datasets and the primary dataset (StepUP-P150) used for the experiments.}
\label{tab:datasets}
\setlength{\tabcolsep}{2pt}
\renewcommand{\arraystretch}{1.15}
\begin{threeparttable}
\begin{tabular}{l l l l l}
\toprule
Dataset & IDs & Strides & Footwear & Other Covariates \\
\midrule
\multicolumn{5}{l}{\textit{Force plate (temporal representations only)}} \\
\midrule
GaitRec \cite{Horsak2020}              & 211 & 7{,}755  & BF (0--1), PF (0--1)               & Speed (3) \\
Gutenberg \cite{Horst2021}             & 350 & 8{,}819  & BF                       & --- \\
Bialystok \cite{DerlatkaParfieniuk2023}& 324 & 13{,}702 & PF (1--7)                & --- \\
ForceID-A \cite{Duncanson2023}         & 184 & 5{,}327  & PF (1--2)                & Speed (3), Days (2) \\
\midrule
\multicolumn{5}{l}{\textit{Pressure platform (spatial representations only)}} \\
\midrule
MFID-200 \cite{Zhou2026}              & 200 & 91{,}796* & BF, SF (4), PF (1)       & Loading (2) \\
\midrule
\multicolumn{5}{l}{\textit{Pressure platform (spatial and temporal representations)}} \\
\midrule
StepUP-P150 \cite{Larracy2025}& 150 & 145{,}084* & BF, SF (1), PF (2)       & Speed (3) \\
\bottomrule
\end{tabular}
\footnotesize{
BF = barefoot, SF = standard footwear (shared across participants), PF = personal footwear. Numbers in brackets denote the number of levels collected per participant. Stride counts marked with {*} include overlapping strides.}
\end{threeparttable}
\end{table}

\subsubsection{Multi-Dataset Training Strategy}

During temporal encoder training, one of the four temporal datasets was randomly selected at each step to act as the auxiliary task, with selection probability weighted by the square root of the dataset's size. 
As there was only one additional spatial dataset (MFID-200), it acted as the auxiliary task for the spatial encoder training at every step.
Batches were drawn from the auxiliary datasets using the same sampling procedure as StepUP-P150 (512 samples from 64 identities, spanning up to four footwear conditions per identity where available, with remaining samples drawn randomly when fewer conditions existed). 
Each auxiliary batch was passed through the model alongside a StepUP-P150 batch, for a combined batch size of 1024. 
Batch normalization was applied jointly across the combined batch, to provide a light data augmentation effect through cross-dataset statistic mixing while improving training stability.

A within-dataset supervised contrastive loss objective was used for training, computed separately for the StepUP-P150 and auxiliary batches to avoid cross-dataset negatives, which could otherwise encourage the model to rely on dataset-specific cues rather than identity-relevant features. 
Under this formulation, the auxiliary dataset is effectively treated as a second, simultaneous identification task, requiring the model to learn representations effective across both datasets concurrently. 
To support convergence under this multi-task setup, we applied PCGrad \cite{Yu2020}. 
When gradients from the two tasks conflicted (i.e., had a negative cosine similarity), each gradient was projected onto the other's normal plane to remove the conflicting components.
This improved optimization across the simultaneous objectives while reducing the influence of dataset-specific patterns on training. 

For the spatial stream specifically, the StepUP-P150 loss was up-weighted by a factor of five before combining it with the MFID-200 loss because of the discrepancy between MFID-200 and StepUP-P150 input representations (due to the channel approximations described above). 
All other training hyperparameters were the same as in single-dataset training (Section \ref{sec:StageI}). 
The two encoders were trained independently with their respective auxiliary dataset(s), then jointly fine-tuned using the StepUP-P150 dataset alone to produce the final encoder.

\section{Results}

\subsection{Part 1: Spatiotemporal Architecture Comparison}

\begin{table}[tb!]
\caption{Identification (rank-1, \%) and verification (EER, \%) performance across three evaluation scenarios.}
\label{tab:model_comparison}
\begin{threeparttable}
\begin{tabular}{@{}lllllll@{}}
\toprule
 & \multicolumn{2}{c}{BF \textrightarrow~BF} & \multicolumn{2}{c}{BF \textrightarrow~SF} & \multicolumn{2}{c}{BF \textrightarrow~PF} \\ 
Model & Rank-1 & EER & Rank-1 & EER & Rank-1 & EER \\ \midrule
R(2+1)D CNN & 96.1 & \textbf{1.9} & 79.2 & \textbf{5.3} & 42.5 & 15.4 \\
3D CNN & 95.8 & 2.3 & 79.2 & 5.9 & 44.6 & 15.5 \\
3D CNN-GRU & 93.3 & 3.0 & 74.3 & 7.0 & 46.3 & 14.6 \\
BiStepNet\textsuperscript{a} & \textbf{97.2} & \textbf{2.0} & \textbf{83.4} & \textbf{4.9} & 52.3 & 12.7 \\
BiStepNet\textsuperscript{b} & 95.9 & 2.4 & \textbf{83.4} & \textbf{5.0} & \textbf{65.2} & \textbf{9.1} \\ \bottomrule
\end{tabular}
\footnotesize
\textsuperscript{a} combined training, \textsuperscript{b} staged training. The best result per column is highlighted in bold; multiple values are bolded where differences were not statistically significant ($p > 0.05$).
\end{threeparttable}
\end{table}

Table~\ref{tab:model_comparison} compares identification and verification performance across the three evaluation scenarios for BiStepNet and the baselines. 
We report macro-averaged rank-1 accuracy (averaged across per-user estimates) and global EER (computed across all users using a single decision threshold). 
Significance for both metrics was assessed using the Wilcoxon signed-rank test on paired per-user estimates across the 150 test users. 

All models performed comparably well within-domain (BF~\textrightarrow~BF), with rank-1 accuracy exceeding 93\% and EER below 3.5\% across the board, indicating that barefoot-to-barefoot recognition is a comparatively easy task for all architectures tested. 
Performance diverged substantially once footwear was introduced at probe time. 
In the BF~\textrightarrow~SF scenario, all models showed a large drop relative to within-footwear performance, but BiStepNet retained the strongest rank-1 accuracy (83.4\%) and lowest EER (4.9\%), outperforming the next-best baselines (3D CNN and R(2+1)D CNN) by more than four percentage points in rank-1 accuracy. 
The BF~\textrightarrow~PF scenario, which introduced unseen footwear domains, proved considerably more difficult for every model, with rank-1 accuracy dropping below 50\% for all baselines. 
Here, while the jointly-trained BiStepNet model still led the baselines (52.3\% rank-1, 12.7\% EER), the staged-training variant achieved 65.2\% rank-1 and 9.1\% EER, a gain of roughly 13 percentage points in rank-1 over the jointly-trained counterpart. 
This improvement in footwear robustness came at a small cost within-domain, where the staged-training model's rank-1 accuracy (95.9\%) was lower than the jointly-trained model's (97.2\%). 

Per-stream evaluation indicated that the improved footwear resilience with the staged training procedure was driven primarily by the temporal stream. 
While the spatial stream performed similarly under both training regimes, the individually trained temporal stream consistently outperformed its end-to-end-trained counterpart, achieving average improvements of 14 percentage points in Rank-1 accuracy and 11 percentage points in EER across the three evaluation scenarios. 
Analysis of the learned gated fusion weights also revealed a higher emphasis on temporal features; across all probe strides, the temporal stream received an average gating weight of 26\% with combined training and 29\% with independent training. 
These findings suggest that staged training enabled the temporal stream to converge to a more discriminative representation, rather than allowing the fusion network to rely predominantly on the stronger, but more footwear-dependent, spatial features. The results indicate that temporal gait information plays an important role in generalization across footwear. The performance of the individual streams is examined in more detail in Section \ref{sec:model_behaviour}.

\subsection{Part 2: Training Set Footwear Representation}

\begin{figure}
    \centering
    \includegraphics[width=\linewidth]{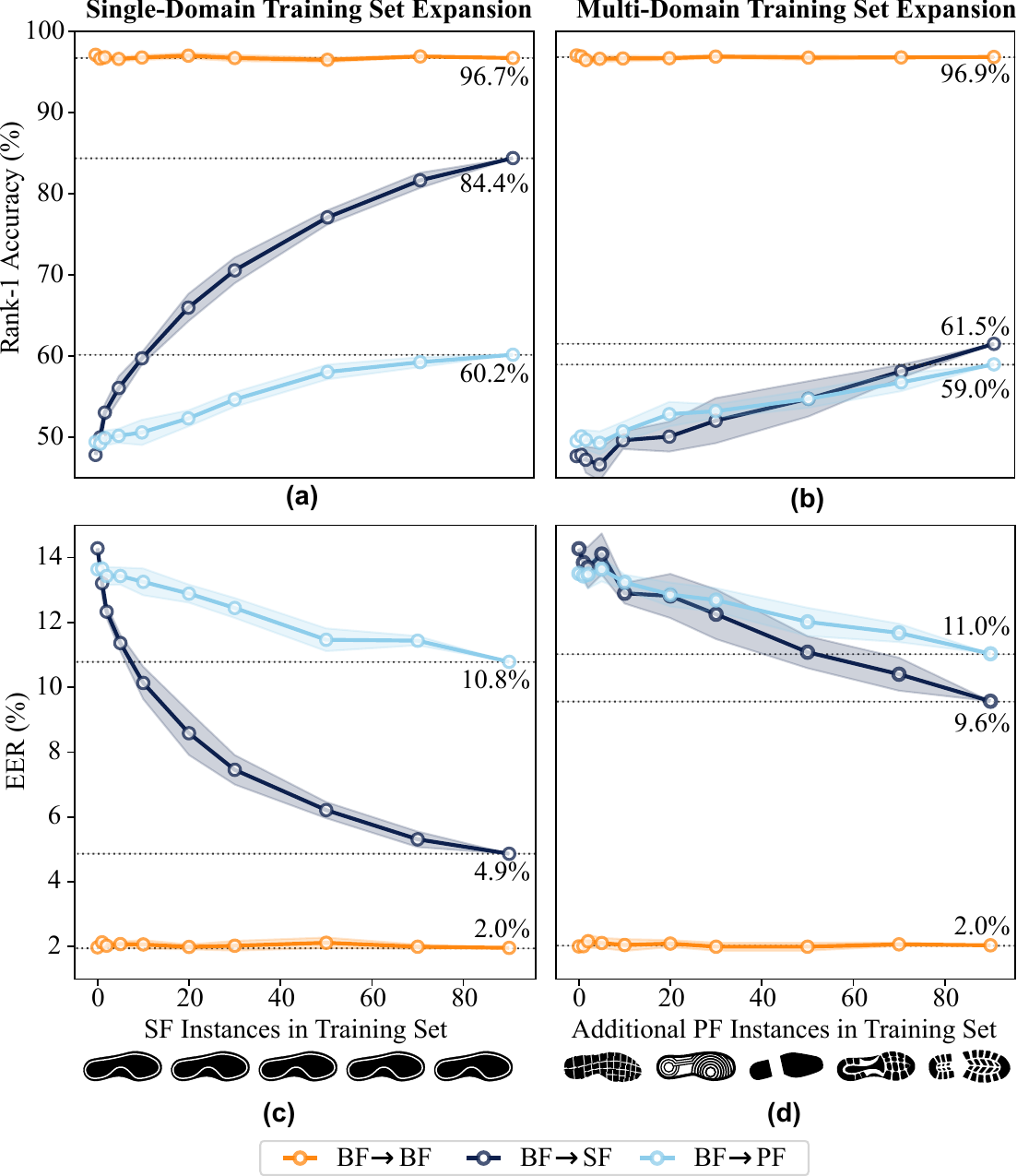}
    \caption{Impact on identification (rank-1) and verification (EER) performance for the three evaluation scenarios as (a,c) the number of standard footwear instances and (b,d) the number of secondary personal footwear instances in the training set are increased. 
}
    \label{fig:learning_curves}
\end{figure}

Fig. \ref{fig:learning_curves} shows identification and verification performance across both training set expansion strategies. 

For single-domain expansion, BF~\textrightarrow~SF performance improved rapidly with minimal representation. As few as two standard footwear instances yielded a significant 5.2 percentage-point improvement in rank-1 accuracy ($p<0.0001$) and a 1.9 percentage-point improvement in EER ($p=0.002$) for new users wearing the same style. 
Gains continued with additional representation, reaching nearly 12 percentage points in rank-1 accuracy and over 4 percentage points in EER by only $k=10$. However, the shape of the curves suggest that BF~\textrightarrow~SF performance had not fully saturated even at $k=90$, remaining more than 10\% below within-domain (BF~\textrightarrow~BF) performance. 

Increasing standard footwear representation also improved performance for personal footwear styles not directly represented by the added instances. 
At $k=90$, gains were observed across all personal footwear categories (Fig. \ref{fig:example_shoes}), including styles visually and structurally dissimilar to the standard shoe. 
The largest relative rank-1 gains occurred for hiking/trail shoes (32.9\%; 14.5 percentage points) and business/dress shoes (32.5\%; 12.9); the largest relative EER gains occurred for flat canvas shoes (33.6\%; 2.6) and boots (30.3\%; 3.2).
BF~\textrightarrow~BF performance remained stable across all values of $k$, indicating that the increased footwear robustness afforded by the added samples did not negatively impact within-domain performance.

As anticipated, standard footwear probe performance was relatively lower under multi-domain expansion (additional personal footwear styles) than under single-domain expansion (standard footwear examples).
However, multi-domain expansion did produce comparable results for BF~\textrightarrow~BF and BF~\textrightarrow~PF evaluations. 
At $k = 90$, the largest relative rank-1 gains were observed for boots (27.9\%; 12.8 percentage points) and sandals (27.6\%; 12.3); the largest relative EER gains were observed for business/dress shoes (34.1\%; 4.9) and flat canvas shoes (29.4\%; 2.3). 
Notably, boots, business/dress shoes, and flat canvas shoes were among the largest gainers under both scaling strategies, suggesting that the categories benefiting most from additional footwear representation were largely consistent regardless of whether that representation came from a single shared style or from diverse, participant-specific styles. 
At $k=90$, there was no significant difference in personal footwear performance between the two training sets, indicating that representing footwear diversity through a single, common shoe style yielded generalization benefits comparable to representing it through varied styles. 

Neither strategy reached saturation in cross-domain personal footwear performance by $k=90$, suggesting that footwear diversity beyond what is feasible to collect from a single participant pool may be needed to fully resolve style-specific shifts; we investigated this using auxiliary datasets in Part 3.




\subsection{Part 3: Multi-Dataset Training}

\begin{table}[tb!]
\caption{Performance improvements using multi-dataset training compared to single-dataset training.}
\label{tab:multi_dataset}
\setlength{\tabcolsep}{5pt}
\begin{threeparttable}
\begin{tabular}{@{}llllllll@{}}
\toprule
 &  & \multicolumn{2}{c}{BF~\textrightarrow~BF} & \multicolumn{2}{c}{BF~\textrightarrow~SF} & \multicolumn{2}{c}{BF~\textrightarrow~PF} \\ 
Encoder & Datasets & Rank-1 & EER & Rank-1 & EER & Rank-1 & EER \\ \midrule
Temporal & Single & 74.9 & 8.1 & 44.3 & 15.4 & 45.2 & 15.5 \\
 & Multi & \textbf{77.0} & \textbf{7.8} & \textbf{46.5} & \textbf{14.6} & \textbf{48.6} & \textbf{14.7} \\ \midrule
Spatial & Single & 95.2 & 2.7 & 75.9 & 6.3 & 42.8 & 15.2 \\
 & Multi & \textbf{96.1} & \textbf{2.4} & 77.0 & 5.7 & 44.1 & \textbf{14.7} \\ \midrule
Fused & Single & 95.9 & 2.4 & 83.4 & 5.0 & 65.2 & 9.1 \\
 & Multi & \textbf{96.5} & \textbf{2.2} & \textbf{85.0} & \textbf{4.7} & \textbf{67.4} & \textbf{8.7} \\ \bottomrule
\end{tabular}
\footnotesize
Bolded values indicate statistically significant improvements compared to single-dataset training ($p < 0.05$).
\end{threeparttable}
\end{table}

Table \ref{tab:multi_dataset} shows performance gains from multi-dataset training for the spatial and temporal encoders individually, and for the fused BiStepNet. Most gains came from the auxiliary temporal datasets, where significant improvements in rank-1 accuracy and EER were observed across all three evaluation scenarios. 
The largest gains occurred for BF~\textrightarrow~PF, with a 3.4 percentage point improvement in rank-1 accuracy and 0.8 in EER. 
Gains were observed across all shoe style categories.
Leave-one-dataset-out analysis confirmed that each auxiliary dataset contributed to the performance, with removal of any single dataset reducing rank-1 accuracy by 23–46\% of the total gain. 
The largest marginal contributions came from the Bialystok and Gutenberg datasets, the two datasets with the most identities by a large margin.  
Notably, while Bialystok includes substantial shoe-style variation (some participants contributing up to seven styles), Gutenberg consists only of barefoot samples, pointing to factors beyond footwear-domain coverage alone as contributing to these gains.

The spatial stream also benefited from multi-dataset training, but to a lesser extent. Rank-1 accuracy improved by 0.9–1.3 percentage points, and EER by 0.3–0.6 across scenarios. 
The improvement was statistically significant within-domain (BF~\textrightarrow~BF: rank-1 $p=0.003$, EER $p<0.0001$), but generally not significant across shoe conditions, aside from a weakly significant EER improvement for BF~\textrightarrow~PF ($p=0.03$). 
This more limited gain may partly reflect the input mismatch between StepUP-P150 and MFID-200, as well as the specific domains represented in MFID-200. For instance, while gains were observed for business/dress shoes and casual sneakers (styles well represented across participants), auxiliary training actually reduced performance for hiking shoes and boots (styles with minimal or no representation). 

Nonetheless, gains from the two streams compounded in the fused spatiotemporal network, with significant improvements across all probe scenarios. 
For the hardest case, BF~\textrightarrow~PF, the fused model achieved a best rank-1 accuracy of 67.4\% and EER of 8.7\%, a 2.2 percentage point improvement in rank-1 accuracy and 0.4 in EER over single-dataset training. 
Rank-1 gains were largest for sandals (6.0 percentage points), followed by business/dress shoes and flat canvas shoes (2.7 each).

\subsection{Characterizing Model Behaviour}\label{sec:model_behaviour}
\begin{figure}[tb!]
    \centering
    \includegraphics[width=\linewidth]{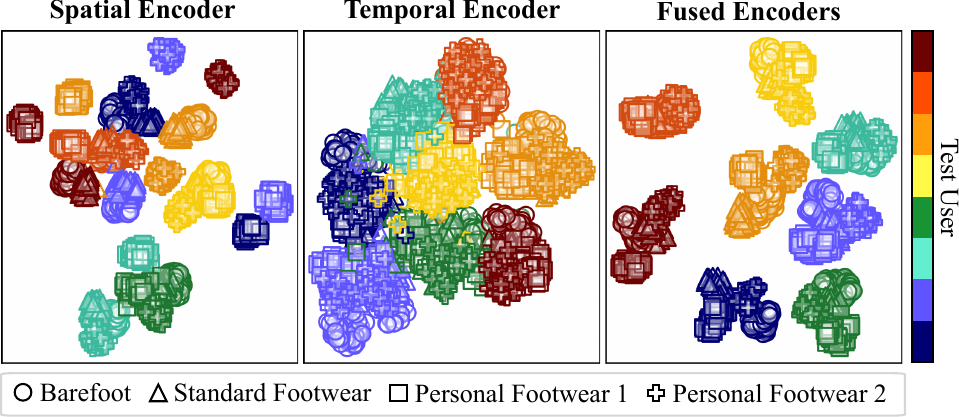}
    \caption{UMAP of the embeddings for eight random test users, from the spatial and temporal encoders and from their fusion. Colors represent different users while marker shapes represent different footwear domains.} 
    \label{fig:example_embeddings}
\end{figure}

Notably, while the spatial encoder outperformed the temporal encoder for well-represented domains (barefoot and standard footwear), the temporal encoder achieved better rank-1 performance for personal footwear (Table \ref{tab:multi_dataset}). 
This corroborates earlier findings on the footwear resilience of temporal features, and helps explain the fused model's approximately 70\% weighting toward spatial embeddings. Spatial features are more identity-discriminative but less resilient to footwear changes, while temporal features trade some discriminative power for cross-shoe robustness.
Figure \ref{fig:example_embeddings} illustrates this through UMAP projections of test-user embeddings from the spatial encoder, temporal encoder, and gated fusion module. 
The spatial encoder produces tight within-condition clusters, but personal footwear samples are frequently displaced from the rest of an identity's cluster, consistent with its lower cross-shoe robustness. 
The temporal encoder shows softer, more overlapping cluster boundaries overall, but keeps a given identity's samples grouped more consistently across footwear conditions. 
The fused model combines these complementary properties, achieving tighter and more unified identity clustering across footwear conditions.

\begin{figure}[tb!]
    \centering
    \includegraphics[width=\linewidth]{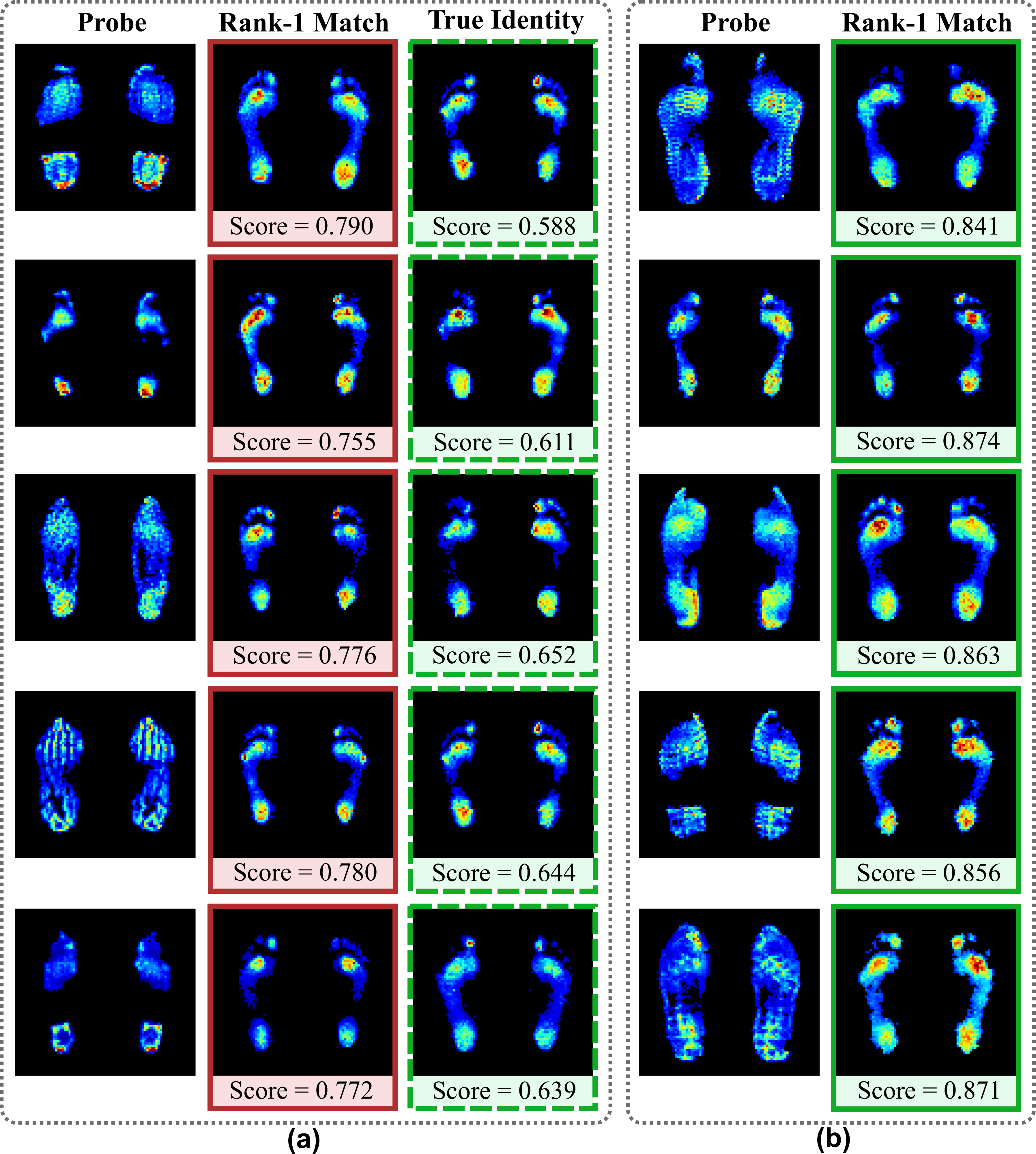}
    \caption{Peak pressure images for extreme probe cases: (a) the most confident false matches (those with the largest similarity score margin over the true identity), and (b) the most confident true matches (largest score margin over the next-ranked identity). }
    \label{fig:example_matches}
\end{figure}

Despite this, several challenging domain shifts could not be overcome by the proposed approach. 
While 31 users achieved BF~\textrightarrow~PF rank-1 accuracy above 90\%, seven fell below 25\% (and three below 10\%). 
Fig.~\ref{fig:example_matches}(a) shows the probes with incorrect rank-1 matches with the largest score margins over the true identity. 
Visually, peak pressure maps are difficult to match across footwear conditions. Although the top-ranked (incorrect) matches often differ from the true identity in characteristics like arch height and foot size, these features are partially obscured by footwear. 
For example, one probe's true identity differs from its rank-1 match by 2 cm in foot length (2.5 UK sizes), yet oversized boots increased the probe's apparent foot length enough to close this gap (row 1).
Conversely, a high heel (row 2) and a sandal (row 3) produced impressions equal to or shorter than their corresponding barefoot impressions. 
Fig.~\ref{fig:example_matches}(b) shows correctly matched probes with the largest score margins over the next-ranked identity; these tend to involve either highly distinctive gait features (e.g., a 12.5 UK shoe size in row 1) or footwear inducing minimal shift (e.g., minimalist shoes designed to mimic barefoot walking in row 2).

\begin{figure}[tb!]
    \centering
    \includegraphics[width=\linewidth]{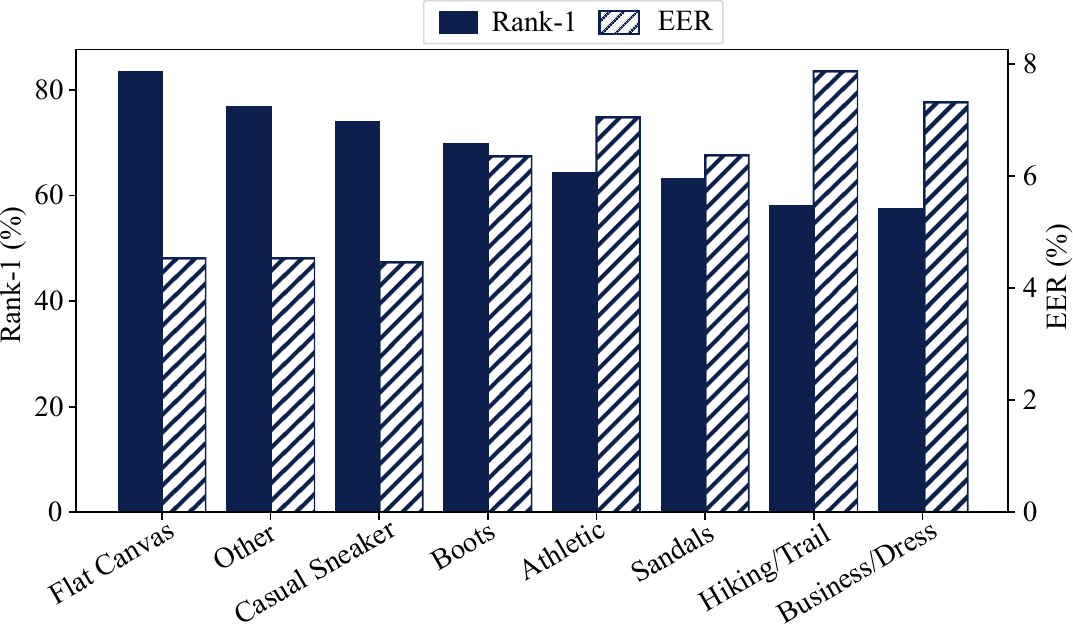}
    \caption{Identification (rank-1) and verification (EER) performance for each shoe style category.}
    \label{fig:pershoe_performance}
    \vspace{-5pt}
\end{figure}

Fig.~\ref{fig:pershoe_performance} shows recognition performance by footwear category. 
Flat canvas shoes and casual sneakers, both relatively homogeneous in sole construction and well represented in StepUP-P150 and the auxiliary datasets, achieved the highest rank-1 accuracy. 
The small ``Other'' category, consisting primarily of minimalist shoes and foam clogs (e.g., Crocs), performed comparably well. 
Business/dress shoes and hiking/trail shoes were the most difficult categories, likely reflecting their greater structural variability and more pronounced gait effects. 
Notably, training set representation alone did not predict performance. Athletic shoes were the most prevalent style in StepUP-P150 (as well as the ForceID-A and Bialystok auxiliary datasets), yet achieved a rank-1 accuracy nearly 20 percentage points lower than flat canvas shoes, plausibly due to the broad structural diversity within the athletic category (running shoes, trainers, basketball sneakers, etc., each with differently structured soles).

\section{Discussion}

\subsection{Implications for Footwear-Robust Representation Learning} 

The proposed BiStepNet architecture significantly outperformed the 3D architecture baselines for cross-shoe recognition, with the largest gains for unseen personal footwear styles (more than 18 percentage points in rank-1 accuracy, 5 in EER), despite the baselines having access to the full 3D footstep volume. 
We attribute this to the inductive biases of the decomposed design, with two streams to independently learn from information-dense spatial and temporal gait representations prior to fusion. 
The architecture is similar in concept to two-stream approaches in both footstep recognition (two parallel ResNets to learn from peak pressure images and pressure sequences separately \cite{CostillaReyes2018}) and action recognition (one stream to encode RGB video frames and another to encode optical flow sequences \cite{Feichtenhofer2016}).

The superior performance of the bi-stream model parallels longstanding findings in fields like action recognition. Architectures that jointly model spatial and temporal structure in a single pathway often carry more parameters and have historically required much larger training sets to match decomposed approaches, only becoming competitive once large-scale datasets such as Sports-1M and Kinetics-400 became available \cite{Alshami2025}. 
Given the current scale of footstep biometric datasets, architectures with stronger structural priors may remain preferable for now, while a shift toward higher-capacity architectures may become viable with additional large datasets.
This trend was already observed in a recent competition that used all 150 users from the StepUP-P150 dataset for model development \cite{Larracy2025first}, where the top solutions used R(2+1)D CNNs. 

Beyond providing a stronger structural prior for limited data, the decomposed architecture may also better preserve temporal information specifically. 
As discussed in Section \ref{sec:model_behaviour}, spatial features are more discriminative in isolation but less footwear-resilient than temporal features. 
When the two streams were trained jointly, spatial features tended to dominate the training signal, consistent with the established phenomenon of modality competition in multi-modal learning \cite{Huang2022}. 
Because the streams are separable, this competition could be addressed directly through staged training, a common strategy in fields such as action recognition \cite{Feichtenhofer2016}, allowing the temporal stream to converge to a more discriminative, footwear-resilient representation than under joint training. 
The 3D architectures, by contrast, have no analogous mechanism to protect weaker but more footwear-robust temporal cues from being overshadowed, a limitation the proposed architecture avoids by design.

\subsection{Learnability of Footwear-Induced Shift} 

Heterogeneous class-conditional shift, in which a domain shift affects different classes in different ways, has been documented in other areas of machine learning such as computer vision.
For instance, a shift from day to night causes pedestrians to be shadowed in darkness while a streetlamp will instead become illuminated \cite{Liu2021}.
A structurally similar problem in biometrics is cross-age face recognition, where the effects of aging can differ across individuals \cite{Zhao2019look}; however, this is generally treated as a source of variability that can be learned and generalized to new users given sufficient training representation. 
We hypothesized that a similar assumption would hold for footwear-induced domain shift.

The results for the standard footwear style provide direct evidence for the learnability and transferability of footwear-robust representations. The trained embeddings transferred well to new users overall (85.0\% rank-1 accuracy), and performance scaled predictably with the number of training identities represented in that footwear style. 
The rapid early gains followed by a long tail toward saturation are consistent with the shift structure identified in Section \ref{sec:problem_formulation}: the dominant, shared component of the domain shift may be learnable from very few examples, while the smaller, participant-specific residual requires substantially more representation to resolve.

This shift structure is also supported by the biomechanics literature. 
The existence of a shared shift component aligns with prior work reporting broadly similar effects across individuals for specific footwear styles, such as high heels \cite{Cronin2014}, flip-flop sandals \cite{Price2014}, and running shoes of varied designs \cite{Lin2022}. 
However, the nature and magnitude of these effects can be modulated by person-specific physiology and behaviour. 
Butler et al. \cite{Butler2006}, for instance, found an interaction between arch type and shoe type during running, with low-arched and high-arched runners exhibiting different loading responses to two sole designs. 
Footwear effects have also been shown to vary with factors like walking speed \cite{Titchenal2015,Xuan2025} and familiarity with a given style \cite{Macdermid2026}. 
Further, Horst et al. \cite{Horst2023id} identified subject-specific responses to different footwear during running.
However, footwear type itself could be classified across participants with high accuracy (median 87.8\%), suggesting that footwear effects are simultaneously person-specific and systematic enough to be learned.

The shape of the learning curve, however, suggests that additional identities are still needed to fully resolve the standard footwear domain. Unlike for personal footwear styles, adding a fourth domain per training user did not improve performance for this domain ($p > 0.05$). 
This suggests that further gains will require additional identities represented in that style (or in styles with similar characteristics), rather than additional domains from existing identities.
Consistent with this, we observed a small but significant increase in BF~\textrightarrow~SF performance when additional identities were incorporated with the auxiliary datasets (Table \ref{tab:multi_dataset}).
Even so, we cannot determine whether the saturated performance for this style would eventually match the high within-domain performance observed for BF~\textrightarrow~BF comparisons. 
It is possible that part of this gap reflects shift components unique to each person that could not be learned even from a much larger training population. 
This question cannot be answered based on the results of the current study.
If there is an unresolvable gap, alternative strategies beyond training set scaling, such as (automated or supervised) reference updating \cite{Pisani2019}, may be needed to close it.

For the more varied personal footwear styles, although there remains substantial room for improvement, we note several promising observations. 
Performance improved both with additional identities represented in existing domains (approximately 10 percentage points in rank-1 accuracy from $k = 0$ to $k = 90$) and with additional domains represented per identity (approximately 50\% rank-1 with two domains, 60\% with three, and 65\% with all four). 
Notably, this scaling held regardless of whether the third domain added per participant was the shared standard shoe or a distinct personal footwear style, indicating that learned features generalize across domains to at least some extent, rather than being specific to the individual style.

This generalization may reflect one or both of two underlying mechanisms. 
First, footwear styles may share a generic barefoot-to-shod shift component, and the model may be learning invariance to this shared component specifically.
This aligns with the average cosine similarity of 0.42 across all barefoot-to-personal-footwear shift vectors that was identified in Section \ref{sec:problem_formulation}, indicating meaningful shared structure even among otherwise dissimilar styles. 
Alternatively, the model may be learning robustness to specific shoe properties, such as the effects of particular heel heights or sole textures, which recur across nominally different styles. 
In this case, adequate training representation may not require direct instances of every specific style, but rather sufficient combined representation of the underlying effects relevant to a style, drawn from a large and varied set of individuals (e.g., the effects of flat soles learned from one group of individuals, and of soft soles from another). 
This suggests that targeted data collection could reasonably prioritize families of styles sharing relevant structural characteristics, such as popular styles or those most representative of a given deployment's expected footwear distribution (e.g., rigid-soled, high-arch business footwear for an office building), rather than attempting to represent every anticipated style directly.

\subsection{Multi-Dataset Training}

The unsaturated learning curves in Part 2 suggest that further gains in cross-shoe recognition require more data than any single available dataset can provide. 
In the absence of a sufficiently large dataset collected under a single protocol, pooling data from multiple sources is a natural alternative. 
Multi-dataset training has proven effective in other domains facing similar data constraints, including lesion detection from medical images \cite{Yan2020}, and semantic segmentation of 3D point clouds \cite{Wu2024}. 
The decomposed architecture adopted in this work, which relies on established, lower-dimensional gait representations, allows each stream to draw on related auxiliary datasets that would be incompatible with a 3D architecture.

However, previous studies have demonstrated the challenges associated with pooling gait datasets from different sources.
Derlatka et al. \cite{Derlatka2024} found that collection location significantly affected ground reaction force measurements despite similar hardware and collection protocols. 
Similarly, Fleischmann et al. \cite{Fleischmann2024} showed that they could predict the source of four different gait datasets with up to 99\% accuracy, indicating that dataset-specific artifacts, rather than genuine gait or identity information, can dominate pooled feature representations if left unaddressed. 
Duncanson et al. \cite{Duncanson2024}, who evaluated combinations of four force plate datasets (including the ForceID-A, GaitRec, and Gutenberg datasets used in this work) reported mixed results on recognition performance. The benefit of the auxiliary datasets depended heavily on dataset compatibility, with many combinations resulting in poorer performance than training with the target dataset alone.

To improve the compatibility and effectiveness of the auxiliary datasets during training, we adopted several strategies.
All samples were resized and normalized to approximately match StepUP-P150, since normalization has been shown to reduce dataset-specific bias in prior gait work \cite{Fleischmann2024}. 
Loss was computed separately within each dataset rather than across pooled batches, to prevent the model from exploiting dataset-specific cues as a shortcut for identity discrimination. 
Finally, PCGrad was used to resolve destructive gradient interference when combining loss gradients from different datasets, encouraging convergence toward a shared solution rather than one that favors either objective at the other's expense. 
Our approach framed multi-dataset training as a form of multi-task learning, in which each dataset constitutes a related task and the model is required to enforce conditional invariance simultaneously across all of them, with every training step informed by both the target and an auxiliary dataset. 

The resulting gains were consistent but modest, likely reflecting the same compatibility limitations observed in prior pooled gait studies. Differences in sensing modality, protocol, and population between StepUP-P150 and the auxiliary datasets constrain how much identity-relevant information can transfer. 
Even with the added datasets, several footwear styles with pronounced or unusual biomechanical effects, such as hiking or winter boots, remained poorly represented in training and correspondingly difficult to recognize. 
Notably, however, gains were not restricted to datasets with rich footwear diversity; Gutenberg, despite consisting entirely of barefoot recordings, contributed a marginal improvement comparable to Bialystok's, whose participants each contributed multiple footwear styles. 
This suggests that at least part of the benefit of auxiliary training comes from reinforcing general, identity-related gait structure, information that helps the model discriminate individuals regardless of footwear, rather than solely from learning explicit footwear-to-footwear shifts.

\subsection{Future Work}
The present study used barefoot reference strides, consistent with previous work \cite{Zhou2026}; barefoot enrollment may be practical in some deployment contexts and provides a common reference domain across users. Moreover, learning the shift between barefoot and shod conditions may be more tractable than between arbitrary pairs of footwear styles. 
Indeed, although the same performance trends were observed when personal footwear styles were used as the reference domain, recognition of other personal styles (PF1~$\rightarrow$~PF2 and PF2~$\rightarrow$~PF1) decreased by an average of 3.4 percentage points in rank-1 accuracy.  
By contrast, using the standard footwear as the reference domain produced an average 7.6 percentage point improvement for personal footwear recognition (SF~$\rightarrow$~PF). 
Thus, the choice of enrollment condition can meaningfully affect cross-footwear performance and should be evaluated according to the requirements of the target application.

The study also assumes five reference strides (10 footsteps, corresponding to approximately two passes across the sensing platform and 10 seconds of walking) and only one probe stride available for each transaction. 
The enrollment requirement is smaller than in several previous studies, including Zhou et al. \cite{Zhou2026} (approximately 28 footsteps), Wang et al. \cite{Wang2024disencb} (72 footsteps), and Connor et al. \cite{Connor2015} (24 footsteps), but larger than the single-stride enrollment assumed by Duncanson et al. \cite{Duncanson2023}. 
Similarly, the present two-footstep transaction provides less information than approaches that aggregate multiple consecutive strides for inference \cite{Connor2015,Zhou2026,Wang2024disencb}. 
Future work should evaluate the trade-off between enrollment effort, transaction duration, and recognition performance.

Lastly, although the dataset includes four walking speeds per participant, the effect of speed was not specifically evaluated in this work. 
All speeds were included in both the development and probe sets, such that most comparisons ($> 75\%$) were cross-speed. 
Thus, some of the observed gains may partially reflect improved robustness to speeds in addition to footwear. 
However, gains from the auxiliary datasets were consistent across speeds ($p=0.70$, Friedman test), while the learning-curve experiments showed either no difference across speeds ($p=0.09$ with single-domain expansion) or greater gains for within-speed comparisons (average 1.5 percentage points higher than cross-speed comparisons with multi-domain expansion; $p = 0.0004$). 
Previous work further suggests that walking speed has a smaller effect on recognition than footwear \cite{Duncanson2023}. 
Nevertheless, explicitly evaluating speed generalization, as well as other deployment-relevant covariates such as load carriage \cite{Derlatka2026}, remains a direction for future work.

\section{Conclusion}
This work investigated whether footwear-robust identity representations can be learned from one population and transferred, zero-shot, to previously unseen users. 
We showed that a bi-stream architecture decomposing spatial and temporal gait representations outperforms 3D baselines for cross-footwear recognition, and that performance scales predictably with the volume and diversity of footwear representation in training, with gains transferring to styles not directly observed in the training set. 
These findings indicate that footwear-induced shift is heterogeneous across individuals, but contains shared, learnable structure.
Nonetheless, performance remains unsaturated, suggesting that more training data than any single available dataset can provide is likely needed to fully resolve it. 
However, we found that auxiliary gait datasets, despite differing sensing modalities and protocols, can be leveraged to improve robustness.

\section*{Acknowledgment}
This work was supported by the New Brunswick Innovation Foundation's Strategic Opportunities Fund, the Atlantic Canada Opportunities Agency's Regional Innovation Ecosystem program, and the Natural Sciences and Engineering Research Council of Canada's Alliance Grants program [ALLRP 558340-20] and Postgraduate Scholarship-Doctoral award [PGS D-589922-2024].

Claude (Anthropic, Sonnet 5) was used to assist with software development and editorial feedback during this work. The authors reviewed and take responsibility for all final content.

\bibliographystyle{ieeetr}
\bibliography{ref}


\begin{IEEEbiography}
[{\includegraphics[width=1in,height=1.25in,clip,keepaspectratio]{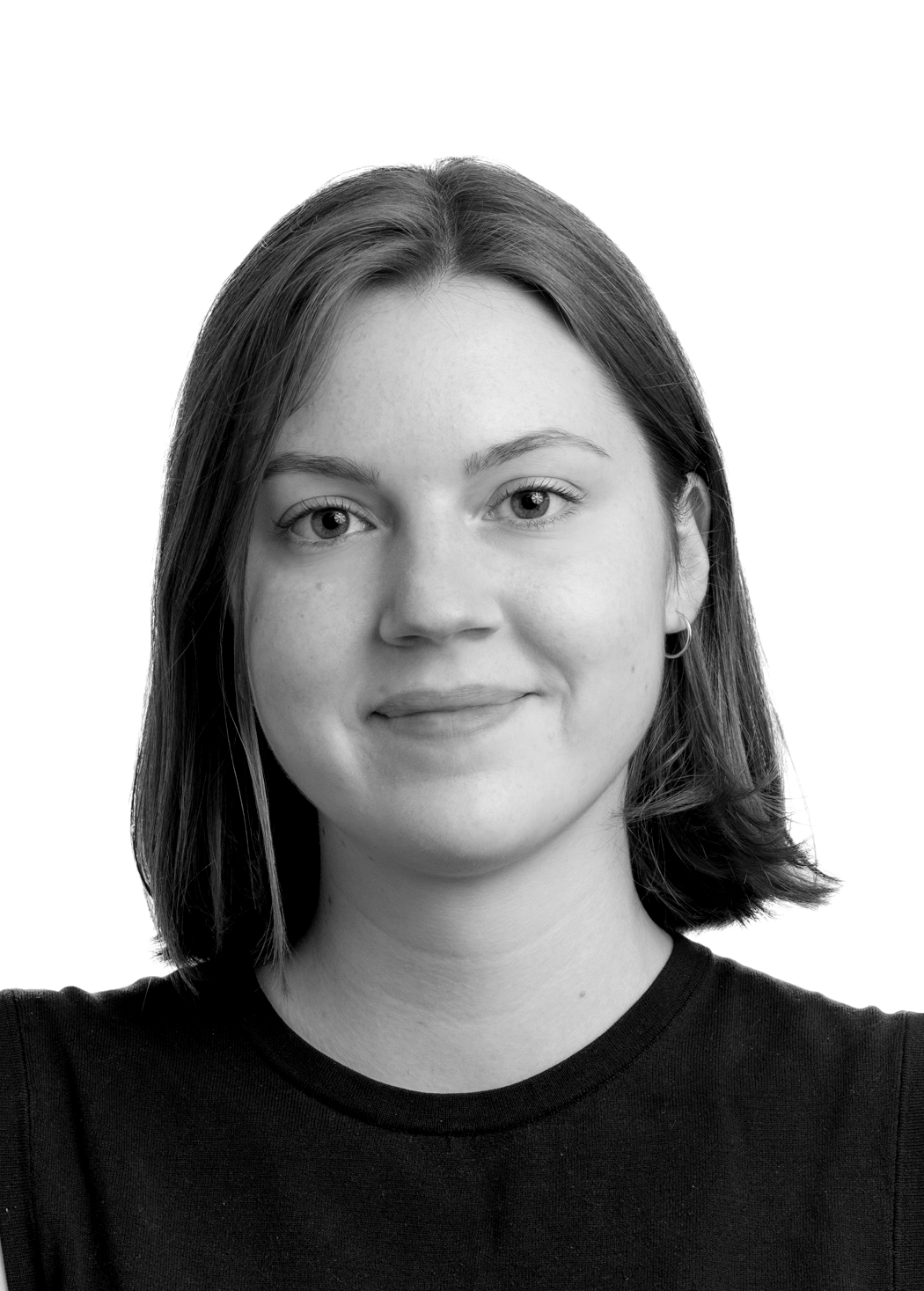}}]{Robyn Larracy}, Graduate Student Member, IEEE, received the B.Sc. and M.Sc. degrees in electrical engineering from the University of New Brunswick, Fredericton, NB, Canada, in 2019 and 2021, respectively, where she is currently pursuing the Ph.D. degree in electrical engineering as a member of MoBIL: the Movement and Biosignals Intelligence Lab. In 2025 and 2026, she hosted the 1st and 2nd International StepUP Competitions for Biometric Footstep Recognition. Her research interests include gait analysis, biometrics, and machine learning for biosignals.
\end{IEEEbiography}

\begin{IEEEbiography}
[{\includegraphics[width=1in,height=1.25in,clip,keepaspectratio]{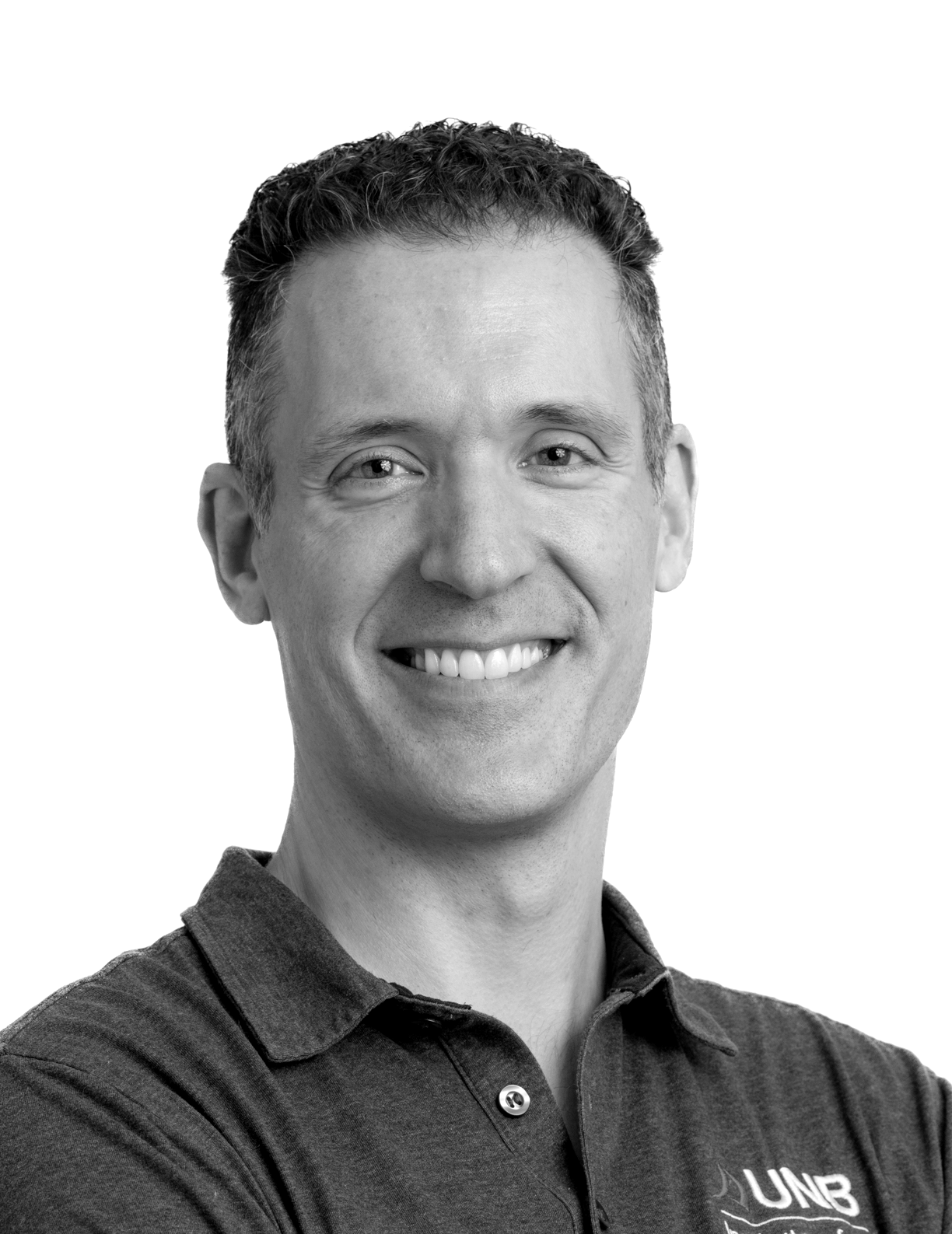}}]{Erik Scheme}, Senior Member, IEEE, received the B.Sc., M.Sc., and Ph.D. degrees in electrical engineering from the University of New Brunswick, Fredericton, NB, Canada, in 2003, 2005, and 2013, respectively. He is a Full Professor with the Department of Electrical and Computer Engineering, the Acting Director of the Institute of Biomedical Engineering, and the Director of MoBIL: the Movement and Biosignals Intelligence Lab at UNB. His research explores the application of signal processing and machine learning for human–machine interfaces, mobility and rehabilitation, digital health, and biometrics. 
\end{IEEEbiography}




\end{document}


\title{Supplementary Material: On the Learnability of Footwear Robustness}
\maketitle

\section{Additional Details for Baseline Networks}

The architectural details for the three spatiotemporal baseline models are in Table \ref{tab:baseline_archs}.
Each of these backbones encode a single 3D footstep pressure recording (101 time steps~\texttimes~75 pixels~\texttimes~40 pixels) into a 256-dim embedding, and the embeddings from the left and right footsteps in a stride are fused using the fully-connected fusion head described in the main text. 

All of the models in this study were implemented with PyTorch 2.10.0 and trained using GPU acceleration with CUDA 12.9.4.
Any convolutional or fully-connected layers were initialized with Kaiming uniform initialization ($a{=}\sqrt{5}$, PyTorch default) \cite{He2015}. 
GRU weight matrices and biases were initialized from $\mathcal{U}(-1/\sqrt{h}, 1/\sqrt{h})$ where $h{=}128$ is the hidden size. 
Affine parameters of all normalization layers were initialized with $\gamma{=}1$, $\beta{=}0$.
PyTorch class implementations for all of the models are available on GitHub\footnote{\href{https://github.com/rlarracy/bistepnet}{github.com/rlarracy/bistepnet}}.

\begin{table}[h!]
\centering
\footnotesize
\setlength{\tabcolsep}{2.8pt}
\caption{
  Architecture details for the spatiotemporal baseline models.
}
\label{tab:baseline_archs}
\begin{threeparttable}
\begin{tabular}{@{}l c c c@{}}
\toprule
\textbf{Layer} & \textbf{Channels} & \textbf{Hyperparameters} & \textbf{Output} \\
\midrule
\multicolumn{4}{@{}l}{\textit{3D CNN (352K params)}}\\
Conv3D$^\dagger$ & 1$\to$32 & $k = (7,5,5)$, $s = (2,2,2)$ & $32{\times}51{\times}38{\times}20$\\
Conv3D$^\dagger$ & 32$\to$64 & $k = (5,3,3)$, $s = (2,2,2)$ & $64{\times}26{\times}19{\times}10$\\
Conv3D$^\dagger$ & 64$\to$128 & $k = (3,3,3)$, $s = (2,2,2)$ & $128{\times}13{\times}10{\times}5$\\
AvgPool3D+FC & 128$\to$256 & pool(1,1,1) & 256\\
\midrule
\multicolumn{4}{@{}l}{\textit{R(2+1)D CNN (364K params)}}\\
Res Conv(2+1)D$^\dagger$ & 1$\to$32 & $k = (7,5,5)$, $s = (2,2,2)$ & $32{\times}51{\times}38{\times}20$\\
Res Conv(2+1)D$^\dagger$ & 32$\to$64 & $k = (5,3,3)$, $s = (2,2,2)$ & $64{\times}26{\times}19{\times}10$\\
Res Conv(2+1)D$^\dagger$ & 64$\to$128 & $k = (3,3,3)$, $s = (2,2,2)$ & $128{\times}13{\times}10{\times}5$\\
AvgPool+FC & 128$\to$256 & pool(1,1,1) & 256\\
\midrule
\multicolumn{4}{@{}l}{\textit{3D CNN-GRU (582K params)}}\\
Conv3D$^\dagger$ & 1$\to$16 & $k = (3,5,5)$, $s = (2,2,2)$ & $16{\times}51{\times}38{\times}20$\\
Conv3D$^\dagger$ & 16$\to$32 & $k = (3,3,3)$, $s = (2,2,2)$ & $32{\times}26{\times}19{\times}10$\\
Conv3D$^\dagger$ & 32$\to$64 & $k = (3,3,3)$, $s = (2,2,2)$ & $64{\times}13{\times}10{\times}5$\\
AvgPool3D+IN & 64$\to$64 & pool(1,1) & $64{\times}13$\\
BiGRU+AttnPool & 64$\to$256 & $h=128$, $l = 2$ & 256\\
FC & 256$\to$256 & -- & 256\\
\bottomrule
\end{tabular}
\footnotesize
IN = InstanceNorm, $k$ = kernel size, $s$ = stride, $h$ = hidden units, $l$ = GRU layers.

$\dagger$ Each is followed by IN and ReLU.
\end{threeparttable}
\end{table}

\section{Auxiliary Dataset Preprocessing}
At the time of writing, all auxiliary datasets used in this work are freely accessible to the research community.

Identical preprocessing was applied to the four force plate datasets to improve consistency during training, following the pipeline used by the GaitRec authors \cite{Horsak2020}. 
Recordings were cropped in time to the interval where vertical GRF exceeded 80 N, filtered with a 2nd-order low-pass Butterworth filter (20 Hz cutoff), and resampled to 101 time steps using nearest-neighbor interpolation. 
For this study, GRF time series were additionally divided by their mean amplitude and COP time series were mean-centered and divided by the total anteroposterior span. 
Strides from the ForceID-A, GaitRec, and Gutenberg datasets were labeled as left or right steps by the original authors. 
For the Bialystok dataset, we applied the authors' provided scripts for automated left/right stride detection \cite{DerlatkaParfieniuk2023}. 

For the MFID-200 dataset, which consists of a single 2D peak pressure map for each trial (capturing approximately 14 consecutive steps on the sensing platform), individual footsteps were extracted using techniques similar to those used for StepUP-P150 \cite{Larracy2025}. 
Pressure maps were thresholded to detect areas of activity, processed using morphological dilation and erosion, and a simple connected-components technique was used to detect the bounding box for each footstep. 
The footstep images were then roughly aligned by rotating the foot's long axis (i.e., its first principal component) to the $y$-axis, downsampled with linear resizing to match the resolution of StepUP-P150 ($4~\rm{px}/cm^2$), and zero-padded to 75~\texttimes~40 pixels. 
During training, pressures above the 99th percentile were clipped, and the resulting maps were min-max normalized and rescaled using a natural log transformation. 
Footsteps were labeled as left or right based on the proportion of active pixels from different quadrants of the forefoot \cite{Macdonald2023}, and consecutive steps were grouped into overlapping left/right pairs to form strides for training.

As with the StepUP-P150 dataset, random spatial affine transformations were applied to the auxiliary datasets during training as a form of augmentation. 
For the time-series datasets, this was implemented through transformation of the mediolateral and anteroposterior COP trajectories:
\begin{align}
\mathrm{ML}' &= \cos(\theta) \cdot \mathrm{ML} - \sin(\theta) \cdot \mathrm{AP} \\
\mathrm{AP}' &= \sin(\theta) \cdot \mathrm{ML} + \cos(\theta) \cdot \mathrm{AP}
\end{align}

\noindent which simulates variations in foot rotation on the force plate. 
Translation was not simulated, since the COP trajectories were mean-centered and therefore unaffected by it.
Likewise, the vertical GRF aggregates forces across the entire foot and is unaffected by spatial changes.

\bibliographystyle{ieeetr}
\bibliography{ref}